\documentclass{article}

     \PassOptionsToPackage{numbers, compress}{natbib}
\usepackage[preprint]{neurips_2021}

\usepackage[utf8]{inputenc} % allow utf-8 input
\usepackage[T1]{fontenc}    % use 8-bit T1 fonts
\usepackage{hyperref}       % hyperlinks
\usepackage{url}            % simple URL typesetting
\usepackage{booktabs}       % professional-quality tables
\usepackage{amsfonts}       % blackboard math symbols
\usepackage{nicefrac}       % compact symbols for 1/2, etc.
\usepackage{microtype}      % microtypography
\usepackage{xcolor}         % colors
\usepackage{amsmath}
\usepackage{graphicx}
\usepackage{wrapfig}

\title{What Is Considered Complete for Visual Recognition?}

\author{%
  Lingxi Xie, Xiaopeng Zhang, Longhui Wei, Jianlong Chang, Qi Tian \\
  Huawei Inc. \\
  \texttt{\{198808xc,zxphistory,weilh2568\}@gmail.com} \\ \texttt{\{jianlong.chang,tian.qi1\}@huawei.com} \\
}

\begin{document}

\maketitle

\begin{abstract}
This is an opinion paper. We hope to deliver a key message that current visual recognition systems are far from complete, \textit{i.e.}, recognizing everything that human can recognize, yet it is very unlikely that the gap can be bridged by continuously increasing human annotations. Based on the observation, we advocate for a new type of pre-training task named \textbf{\textit{learning-by-compression}}. The computational models (\textit{e.g.}, a deep network) are optimized to represent the visual data using compact features, and the features preserve the ability to recover the original data. Semantic annotations, when available, play the role of weak supervision. An important yet challenging issue is the evaluation of image recovery, where we suggest some design principles and future research directions. We hope our proposal can inspire the community to pursue the compression-recovery tradeoff rather than the accuracy-complexity tradeoff.
\end{abstract}

\section{Introduction}
\label{introduction}

Recognition plays a central role in computer vision, where the goal is to locate and depict the basic elements in an image, a video clip, or even more complicated data stream. In the deep learning era~\cite{lecun2015deep}, the convolutional neural networks have been widely applied for various kinds of recognition tasks including image-level classification~\cite{deng2009imagenet}, object-level detection and segmentation~\cite{everingham2010pascal,lin2014microsoft}, part-level understanding (\textit{e.g.}, human pose estimation~\cite{andriluka20142d}), \textit{etc}. The ability of visual recognition has also boosted other vision problems such as low-level vision (\textit{e.g.}, super-resolution~\cite{agustsson2017ntire}) and multi-modal understanding (\textit{e.g.}, image captioning~\cite{lin2014microsoft} and text-to-image retrieval~\cite{young2014image}).

\begin{figure}[!b]
\centering
\includegraphics[width=1.00\textwidth]{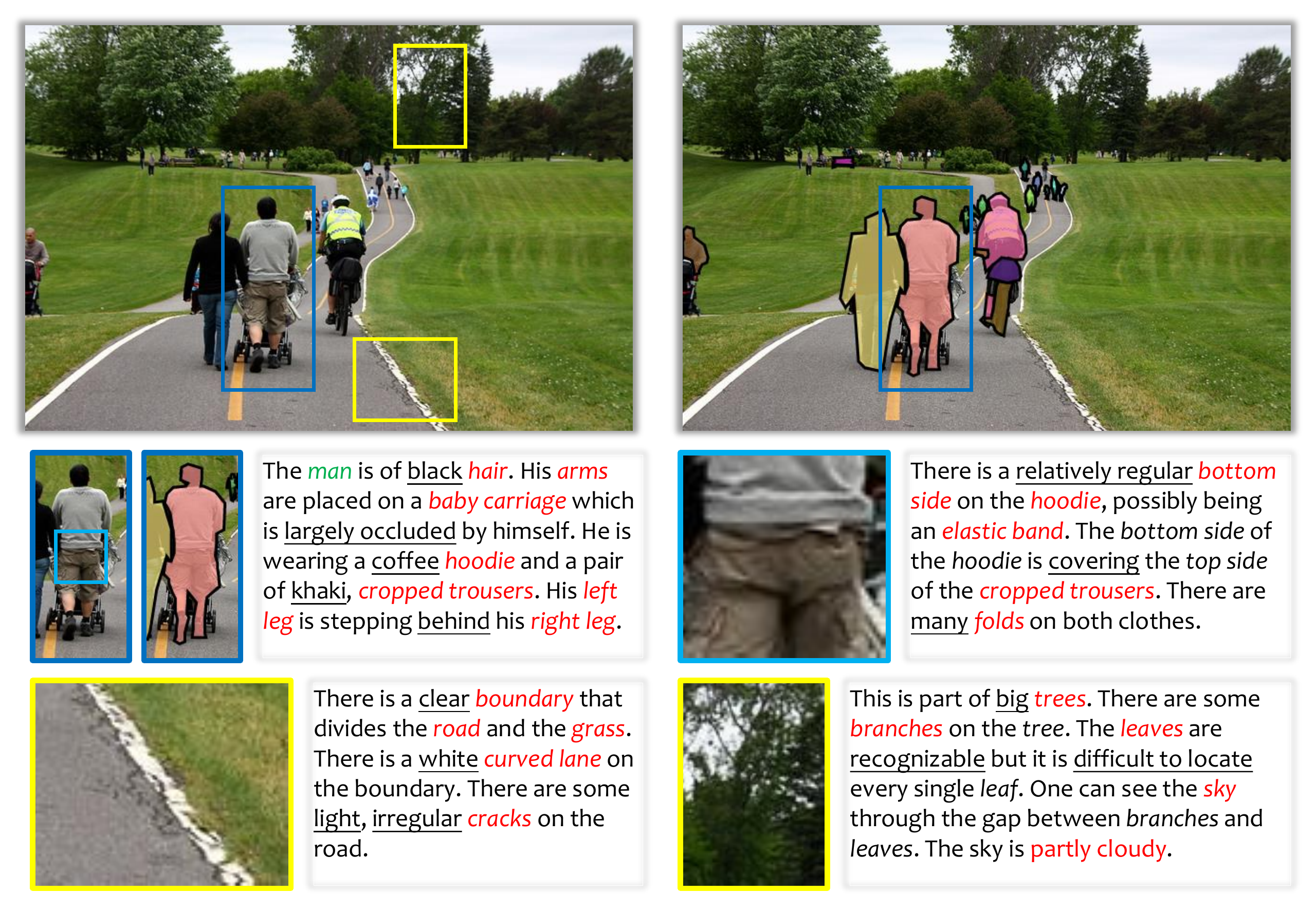}
\caption{The motivation of this work, \textit{i.e.} human can easily recognize much richer contents than the available annotations. The top row shows an image from the MS-COCO dataset~\cite{lin2014microsoft} and the corresponding instance-level segmentation. The \textcolor{blue}{blue} box indicates an instance that is holistically annotated as a \textit{person}, but there are missing details such as his \textit{arms}, \textit{legs}, and \textit{clothes}. Going one step further, the \textcolor{cyan}{cyan} box contains a part of human body and the \textcolor{yellow}{yellow} boxes indicate two regions which are not considered instances. All these regions contain rich recognizable details but they are often hard to annotate. We describe these details in texts, but even the texts do not cover everything recognizable. The \textit{italics} denote concepts and the \underline{underlines} denote attributes. The \textcolor{green}{green} and \textcolor{red}{red} texts indicate \textcolor{green}{labeled} and \textcolor{red}{unlabeled} concepts, respectively. \textbf{Even provided sufficient human labors, it is unlikely that everything can be annotated.} This figure is best viewed in color.}
\label{fig:motivation}
\end{figure}

Despite the remarkable advance of visual recognition in the past decade, researchers have noticed the limitations of deep networks~\cite{goodfellow2015explaining,samek2017explainable}, in particular, `the fundamental difficulties to deal with the enormous complexity of natural images'~\cite{yuille2021deep}. In this paper, we put forward an important problem: \textcolor{blue}{\textbf{What is considered \textit{complete} for visual recognition?}} Here by `complete' we refer to a vision algorithm that recognizes all visual information that human can recognize. However, as shown in Figure~\ref{fig:motivation}, this goal is far from complete. The existing datasets often offer basic visual information (\textit{e.g.}, independent object instances like a \textit{person} and uncountable concepts like the \textit{sky}), but obviously, human can easily recognize rich and hierarchical information upon request. For example, the class of \textit{person} appears in many datasets, but few datasets have labeled the parts (\textit{head}, \textit{arms}, \textit{body}, \textit{etc.}) of each \textit{person}. Even with \textit{arms} labeled, human can further recognizing \textit{hands}, \textit{fingers}, \textit{knuckles}, and even \textit{fingerprints}, if he/she desires. In addition, the level of supervision varies significantly across different scenarios (\textit{e.g.}, in the MS-COCO dataset, \textit{persons} are assigned with different instance labels but other objects like \textit{buildings} and \textit{trees} are not.

That is to say, even with unlimited human labor, it is unlikely that the obtained information can be made complete. The limited granularity of annotation obstructs the development of visual recognition, in particular, the current \textbf{\textit{learning-by-annotation}} paradigm. This drives us to design a new paradigm for visual recognition. Inspired by the conjecture that human tends to perceive complex data by finding the easiest way to understand it~\cite{rissanen1978modeling,grunwald2007minimum}, we propose a new task named \textbf{\textit{learning-by-compression}}. It evaluates a computational model by the ability of compressing image/video data into compact features and restoring the visual contents using the compressed features. Mathematically, provided an image $\mathbf{x}$, there is an encoder $f\!\left(\cdot\right)$ that represents it into a vector $f\!\left(\mathbf{x}\right)$ of minimal length, meanwhile, there exists a decoder $g\!\left(\cdot\right)$ that restores the image using $f\!\left(\mathbf{x}\right)$. The restoration cannot be perfect, and a function $r\!\left(\cdot,\cdot\right)$ is needed to evaluate the recovery quality of $\mathbf{x}'$ with respect to $\mathbf{x}$.

The key component of learning-by-compression is the evaluation function, \textit{i.e.}, $r\!\left(\cdot,\cdot\right)$. We conjecture that the design of $r\!\left(\cdot,\cdot\right)$ offers an alternative path to delve into the core difficulty of computer vision, yet we do not think that the functions simply based on pixel statistics, such as PSNR and SSIM~\cite{wang2004image}, can work well. Once a reasonable $r\!\left(\cdot,\cdot\right)$ is found, the learning objective is to adjust a pair of $f\!\left(\cdot\right)$ and $g\!\left(\cdot\right)$ to arrive at a better tradeoff between data compression and image recovery, which we believe is more worth pursuing than the tradeoff between recognition accuracy and model size (see Figure~\ref{fig:tradeoff}). Learning-by-compression can be understood as a new visual pre-training task in which we do not assume the existence of semantic labels while the labels can assist the algorithm to achieve a higher compression ratio. The pre-trained encoder and decoder offer a more flexible usage in downstream tasks, covering both high-level and low-level visual understanding, which stands out from existing visual pre-training algorithms.

The remaining part of this paper is organized as follows. Section~\ref{difficulty} argues that incomplete annotations may have obstructed the development of computer vision. Section~\ref{proposal} presents our proposal that formulates vision pre-training as data compression. After discussing the relationship to prior work in Section~\ref{related}, we offer a brief summary in Section~\ref{summary}.

\section{Visual Recognition is Obstructed by Incomplete Annotations}
\label{difficulty}

Our goal is to investigate the \textbf{recognizable semantics} of an image, by which we refer to the information that a regular labeler can recognize if he/she wants. Also, we make an assumption that for natural images, the recognizable semantics are mostly the same across different labelers. Consider the example in Figure~\ref{fig:motivation}. This is a regular picture that, by the standard of the MS-COCO dataset~\cite{lin2014microsoft}, has an intermediate level of complexity. The objects that are defined by MS-COCO (\textit{e.g.}, \textit{persons}, \textit{bikes}, \textit{etc.}) have been pixel-wise annotated, but the annotation is far from complete in at least three aspects, namely, (i) the objects can be divided into parts (\textit{e.g.}, the \textit{head} and \textit{arm} of a \textit{person}), (ii) there are rich attributes for these objects (\textit{e.g.}, \textit{color} or \textit{texture}), and (iii) there are objects that do not appear in the MS-COCO dictionary (\textit{e.g.}, the \textit{cracks} on the \textit{road}, the \textit{leaves} of a \textit{tree}).

Generally speaking, the visual concepts are often organized in a hierarchical way. The currently popular datasets often annotated the top-level concepts but ignored the parts and/or attributes affiliated to the objects. On the other hand, even provided sufficient time and resource, it is unlikely that a well-defined annotation system can be established to support all recognizable visual concepts because (i) the recognizable concepts form an open set~\cite{scheirer2012toward,hu2018learning} and, more importantly, (ii) there is always a tradeoff between the granularity and accuracy of annotation. The second point embodies in different aspects, including object classification (basic object categories are easy to distinguish, but the boundary becomes ambiguous for finer-level categories) and segmentation (as annotation goes to the bottom level, the boundary between neighboring concepts becomes less clear while the impact of pixels or even sub-pixels becomes more significant).

Currently, most vision algorithms follow the paradigm of \textbf{\textit{learning-by-annotation}}. As the recognition accuracy on standard benchmarks goes up, the incomplete annotations raise a serious issue that the algorithms may have learned biased data. For example, to achieve satisfying detection and segmentation accuracy in a high IOU threshold, the object borders need to be accurately depicted, but assigning the same label for all boundary pixels is often inaccurate. When the amount data is limited, the algorithms can be biased towards over-fitting the training set and even the evaluation protocol built upon incomplete annotations. This is part of the reason that downgrades the transfer performance of vision algorithms -- there are side evidences that richer annotations are useful for visual recognition, \textit{e.g.}, bounding-box annotations improves image classification~\cite{angelova2013efficient}, and pixel-level annotations improves object detection~\cite{he2017mask}.

To bypass the difficulty of incomplete annotations, we propose a new paradigm named \textbf{\textit{learning-by-compression}} that, instead of recognizing specific concepts, pursues the ability to represent an image using minimal information that can be used to recover the major contents of the image. The idea behind learning-by-compression is that a computational model must learn an efficient vocabulary from a large-scale dataset so that the visual contents, labeled or unlabeled, can be encoded into compact features. We describe the proposal in the following sections.

\section{Visual Representation Learning as Data Compression}
\label{proposal}

\begin{figure}[!t]
\centering
\includegraphics[width=1.00\textwidth]{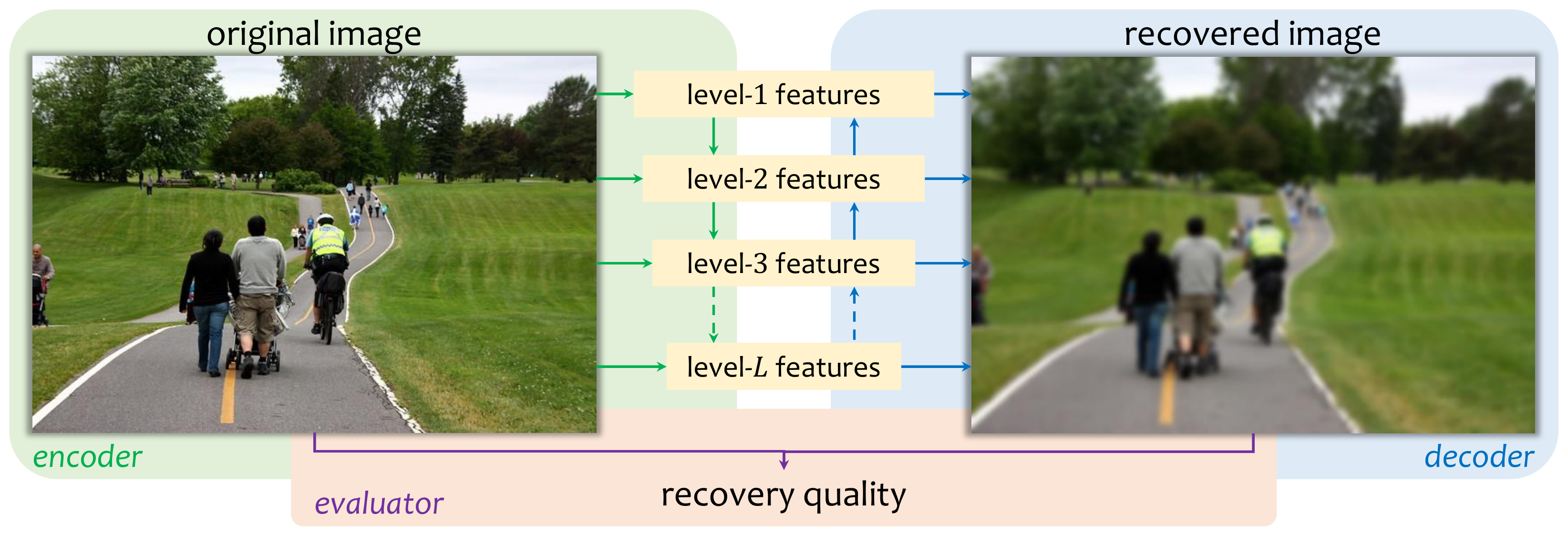}
\caption{The conceived framework of self-supervised representation learning driven by data compression. For brevity, we show a layer-wise deep network that uses multiple levels of features to represent the image, while the computational model and the features can be in arbitrary forms. It is the tradeoff between compression and recovery that really matters.}
\label{fig:framework}
\end{figure}

We formulate our idea into a vision pre-training framework. Mathematically, let $\mathbf{x}$ denote a training image, where we do not assume any type of annotation to exist, yet we believe that supervision can boost visual representation learning (see Section~\ref{proposal:optimization}). The goal is to represent $\mathbf{x}$ using a compact feature vector, $\mathbf{z}$, that can be used to recover the original image. The overall framework is illustrated in Figure~\ref{fig:framework}. There are three key components in the proposed framework, namely, an \textbf{\textit{encoder}} $f\!\left(\cdot\right)$ that compresses $\mathbf{x}$ into $\mathbf{z}$, a \textbf{\textit{decoder}} $g\!\left(\cdot\right)$ that generates $\mathbf{x}'$ from $\mathbf{z}$, and an \textbf{\textit{evaluator}} $r\!\left(\cdot,\cdot\right)$ that measures the difference between the recovered image and the original image.

\begin{figure}[!t]
\centering
\includegraphics[height=5cm]{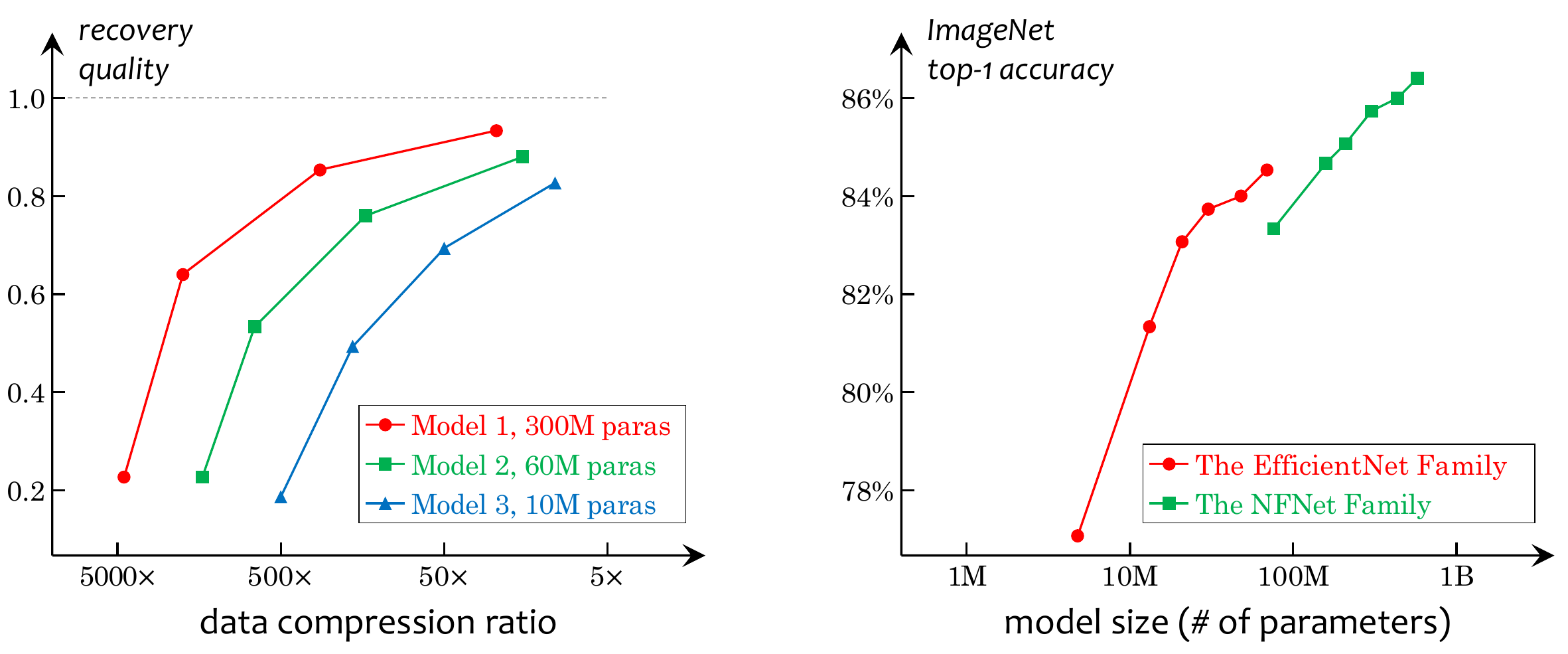}
\caption{\textbf{Left:} an imaginary example of the tradeoff between data compression and image recovery. Three broken lines indicate three models with different sizes, where we believe that increasing the model size can facilitate a better tradeoff. \textbf{Right:} for comparison, we show how the state-of-the-art visual recognition models~\cite{tan2019efficientnet,brock2021high} pursue a tradeoff between model size and accuracy.}
\label{fig:tradeoff}
\end{figure}

We assume that a dataset $\mathcal{D}$ exists for training the models, $f\!\left(\cdot\right)$ and $g\!\left(\cdot\right)$. For any pair of $f\!\left(\cdot\right)$ and $g\!\left(\cdot\right)$, we can compute the data compression ratio and the recovery quality, both averaged on $\mathcal{D}$:
\begin{equation}
\label{eqn:criterion}
{\overline{\mathrm{CR}}}= {\frac{\sum_{\mathbf{x}\in\mathcal{D}}\mathrm{size}\!\left(\mathbf{x}\right)}{\sum_{\mathbf{x}\in\mathcal{D}}\mathrm{size}\!\left(f\!\left(\mathbf{x}\right)\right)}},\quad\quad\quad\quad{\overline{\mathrm{RQ}}}={\frac{1}{\left|\mathcal{D}\right|}\sum_{\mathbf{x}\in\mathcal{D}}-\log r\!\left(\mathbf{x},g\!\left(f\!\left(\mathbf{x}\right)\right)\right)}.
\end{equation}
Here, we use $\mathrm{size}\!\left(\cdot\right)$ to denote the data size, \textit{i.e.}, the number of elements in the original and compressed data. Obviously, there is a tradeoff between $\overline{\mathrm{CR}}$ and $\overline{\mathrm{RQ}}$, \textit{i.e.}, as the compression ratio goes up, the recovery quality is deemed to drop. Hence, the goal of pre-training is to push the tradeoff towards the goal that highly-compressed feature vectors can still generate high-quality images.

As a side note, the reason that an image can be compressed is that the computational model (\textit{e.g.}, the combination of deep networks, $f\!\left(\cdot\right)$ and $g\!\left(\cdot\right)$) has stored efficient visual patterns. Once well optimized, a computational model with more parameters has a stronger abilities of visual representation, so that a better tradeoff between compression and recovery can be achieved, as shown in Figure~\ref{fig:tradeoff}. This offers a new perspective of the emerging trend of large-scale pre-trained models. Note that each computational model can produce different versions of compression and recovery by inserting lightweight projection functions between $f\!\left(\cdot\right)$ and $g\!\left(\cdot\right)$, which we shall elaborate in Section~\ref{proposal:architecture}.

In what follows, we discuss the design principles of the individual modules including the encoder, decoder, and, more importantly, the evaluator that measures the quality of image recovery.

\subsection{The Encoder-Decoder Architecture}
\label{proposal:architecture}

We start with the state-of-the-art design that the encoder and decoder, $f\!\left(\cdot\right)$ and $g\!\left(\cdot\right)$, are deep networks\footnote{Throughout this paper, we assume that $f\!\left(\cdot\right)$ and $g\!\left(\cdot\right)$ are convolutional networks, yet the framework does not obstruct using Transformers~\cite{dosovitskiy2021image,touvron2020training} as the computational model.} that contain a number of layers. According to the common expertise, different layers tend to respond to different visual information -- the high-resolution layers (close to the image) are sensitive to low-level patterns (\textit{e.g.}, texture, geometry), while the low-resolution layers can capture high-level semantics with fewer details. Therefore, to achieve a good tradeoff between compression and recovery, both $f\!\left(\cdot\right)$ and $g\!\left(\cdot\right)$ should integrate multi-level information.

We first elaborate the design of $f\!\left(\cdot\right)$, and the reverse design derives $g\!\left(\cdot\right)$. Denote the input layer as ${\mathbf{x}_0}\equiv{\mathbf{x}}$. Let there be $L$ modules for hierarchical feature extraction. For ${l}={1,2,\ldots,L}$, $\mathbf{x}_l$ indicates the output of the corresponding module, $f_l\!\left(\cdot\right)$, \textit{i.e.}, ${\mathbf{x}_l}={f_l\!\left(\mathbf{x}_{l-1}\right)}$, which has a spatial resolution of $W_l\times H_l$ and has $C_l$ channels, \textit{i.e.}, ${\mathrm{size}\!\left(\mathbf{x}_l\right)}={W_l\times H_l\times C_l}$. Note that $f_l\!\left(\cdot\right)$ can be a single layer or an arbitrarily complex module, \textit{e.g.}, an intermediate stage in ResNet~\cite{he2016deep}, DenseNet~\cite{huang2017densely}, or a module obtained by neural architecture search~\cite{zoph2018learning}.

The compact representation of $\mathbf{x}$ comes from $\left\{\mathbf{x}_l\right\}_{l=1}^L$. To facilitate data compression, an auxiliary module, $f_l'\!\left(\cdot\right)$, is inserted to convert $\mathbf{x}_l$ into $\mathbf{z}_l$, ${l}={1,2,\ldots,L}$, with the dimensionality reduced. We name the size of elements in $\left\{\mathbf{x}_l\right\}_{l=1}^L$ and $\left\{\mathbf{z}_l\right\}_{l=1}^L$ as the original and compressed configurations of feature extraction. For each encoder, $f\!\left(\cdot\right)$, the original configuration remains constant and will be used in downstream tasks, while the compressed configuration can be adjusted (an example is shown in Table~\ref{tab:configuration}) to produce a tradeoff between data compression and recovery quality (see Figure~\ref{fig:tradeoff}). The design of $g\!\left(\cdot\right)$ is straightforward. The compressed features, $\left\{\mathbf{z}_l\right\}_{l=1}^L$, are fed into another set of auxiliary modules for dimensionality growth. The intermediate outputs are then formulated as ${\mathbf{x}_L'}={g_L'\!\left(\mathbf{z}_L\right)}$, ${\mathbf{x}_l'}={g_l'\!\left(\mathbf{z}_l\right)+g_{l+1}\!\left(\mathbf{x}_{l+1}\right)}$ for ${l}={L-1,\ldots,2,1}$, and ${\mathbf{x}'}\equiv{\mathbf{x}_0'}={g_1\!\left(\mathbf{z}_1\right)}$.

We leave two side notes here. First, the ultimate goal of pre-training is to improve visual recognition, hence we need to consider how the encoder-decoder architecture is inherited by downstream tasks. We shall delve into details in Section~\ref{proposal:application}. Here, we emphasize that if the downstream task involves only the encoder (\textit{e.g.}, image classification or object detection) or decoder (\textit{e.g.}, conditional image generation), the auxiliary modules may not be inherited. To be friendly to these scenarios, the auxiliary modules shall not be very complicated, \textit{e.g.}, formulated as one or two linear projection layers. Second, the proposed architecture involves both encoder and decoder, which stands out from the state-of-the-art self-supervised learning proxies~\cite{oord2018representation,grill2020bootstrap}. As we shall see in Section~\ref{related:learning}, our algorithm is easily specialized into these proxies.

\begin{table}[!t]
\centering
\begin{tabular}{c|c|c|ccc}
\hline
Module \# & Resolution & Original Image & Variant \#1 & Variant \#2 & Variant \#3 \\
\hline
0 & $256\times256$ & $3$ & $0$ & $0$ & $0$ \\
1 & $64\times64$ & $0$ & $2$ & $1$ & $0$ \\
2 & $32\times32$ & $0$ & $4$ & $2$ & $2$ \\
3 & $16\times16$ & $0$ & $8$ & $4$ & $3$ \\
4 & $8\times8$ & $0$ & $16$ & $8$ & $4$ \\
\hline
$\mathrm{CR}$ & $-$ & $1.0\times$ & $12.8\times$ & $25.6\times$ & $64.0\times$ \\
\hline
\end{tabular}
\caption{An example of the original network and three compressed configuration. The input is an RGB image of $256\times256$. Each grid shows the dimensionality (spatial resolution and channel number) of an intermediate output. Feature compression may change the spatial resolution (\textit{e.g.}, by $2\times2$ convolution) and/or channel number (\textit{e.g.}, by linear projection). Both the original and compressed features can be further compressed using JPEG or other algorithms -- for brevity, we compute the compression ratio using the length of all features (\textit{i.e.}, as the raw data are directly stored).}
\label{tab:configuration}
\end{table}

\begin{wrapfigure}{r}{6cm}
\vspace{-0.5cm}
\includegraphics[width=6cm]{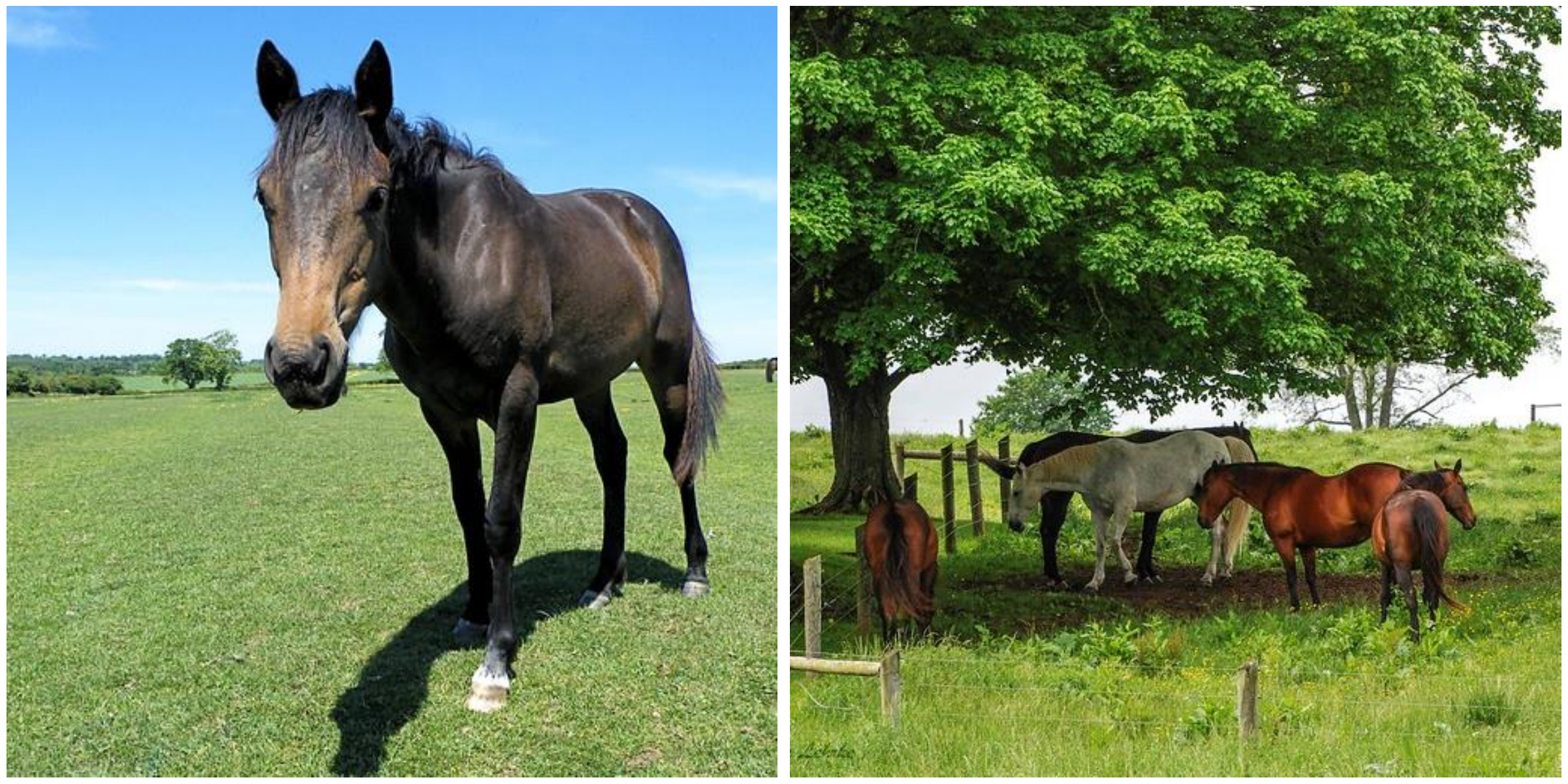}
\caption{Two \textit{horse} images from MS-COCO~\cite{lin2014microsoft}, where the left one contains less information than the right one.}
\label{fig:examples}
\vspace{-1.0cm}
\end{wrapfigure}

Last but not least, we point out that the compression ratio does not need to be constant for all images. As shown in Figure~\ref{fig:examples}, it is possible to use a shorter vector to represent an image with simple geometry and clean background, while a longer vector to represent a web image with rich details. This can be done by the dynamic routing networks~\cite{larsson2017fractalnet,huang2018multi} that incrementally add new modules until the stored information is sufficient for image recovery. The average compression ratio is computed on a sufficiently large dataset that is believed to capture the data distribution of interest.

\subsection{The Evaluation of Image Recovery}
\label{proposal:evaluation}

\begin{figure}[!t]
\centering
\includegraphics[width=0.98\textwidth]{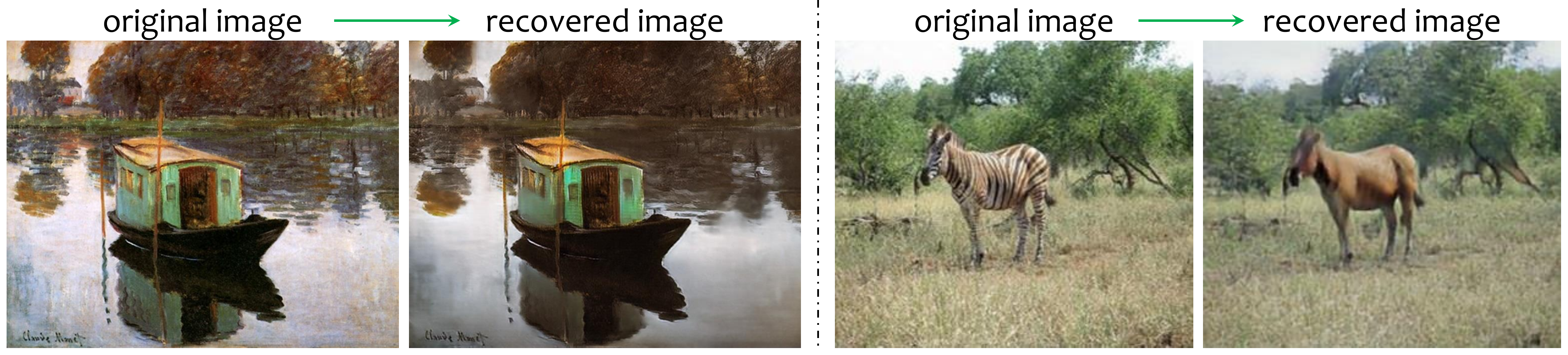}
\caption{Two pairs of original and recovered images. We have borrowed the examples from CycleGAN~\cite{zhu2017unpaired} for illustration. \textbf{Left:} the algorithm converts the image style from an oil painting into a real photo -- while the intensity of many pixels has been largely modified, the main contents are restored and thus the evaluator should produce a high score. \textbf{Right:} the algorithm converts the foreground from a \textit{zebra} to a \textit{horse} -- oppositely, the evaluator shall produce a low score though most pixels (the background) remain unchanged.}
\label{fig:evaluation}
\end{figure}

We aim to deliver the key message: \textbf{evaluating the recovery quality is a difficult task.} This can be easily realized using the examples in Figure~\ref{fig:evaluation}. Also, imagine another input image with a \textit{zebra} is standing on the \textit{grass}, and the recovered image may change the texture of either the \textit{zebra} or the \textit{grass}. A good evaluator should focus on the \textit{zebra} rather than the \textit{grass} because the \textit{zebra} is semantically important. That is to say, the recovery score should be significantly reduced if the \textit{stripes} of \textit{zebra} are contaminated, but remain mostly unaffected if the texture of \textit{grass} is perturbed, although both cases lead to a similar drop of pixel-level similarity.

Let us formulate the evaluator $r\!\left(\cdot,\cdot\right)$, the key component of $\overline{\mathrm{RQ}}$ (see Eqn~\eqref{eqn:criterion}), as
\begin{equation}
\label{eqn:evaluator}
{r\!\left(\mathbf{x},\mathbf{x}'\right)}={\mathbf{w}\!\left(\mathbf{x},\mathbf{x}'\right)^\top\cdot\left|p\!\left(\mathbf{x}\right)-p'\!\left(\mathbf{x}'\right)\right|},
\end{equation}
where $p\!\left(\cdot\right)$ and $p'\!\left(\cdot\right)$ are pre-defined proxy functions that extract semantically important information from $\mathbf{x}$ and $\mathbf{x}'$, respectively, and $\mathbf{w}\!\left(\cdot,\cdot\right)$ denotes the function that adds weights to the terms. Eqn~\eqref{eqn:evaluator} is a generalized form. When ${p\!\left(\mathbf{x}\right)}\equiv{\mathbf{x}}$, ${p'\!\left(\mathbf{x}'\right)}\equiv{\mathbf{x}'}$, and ${\mathbf{w}\!\left(\mathbf{x},\mathbf{x}'\right)}\equiv{\mathbf{1}}$, then $r\!\left(\mathbf{x},\mathbf{x}'\right)$ degenerates to pixel-wise similarity, and is closely related to the popular metrics of PSNR and SSIM~\cite{wang2004image}. However, this is not a good choice due to the example in Figure~\ref{fig:evaluation}. From another perspective, this design results in that the algorithm devotes part of information to represent less important image data, which eventually downgrades the recognition accuracy in downstream tasks.

We suggest another solution that involves (i) semantic supervision on part of training data and (ii) gradual adjustment of weights. For each training sample $\mathbf{x}$ and the corresponding recovered sample $\mathbf{x}'$, we apply the same proxy $p\!\left(\cdot\right)$ that is composed of two parts, with the first part extracting patch-wise statistics (\textit{e.g.}, patch-level mean and variance) and the second part performing semantic recognition (\textit{e.g.}, object detection, according to available supervision). In a generalized learning environment, supervision may be absent on many training samples, in which case the second part is considered all-zero. Therefore, the proposal is understood as weakly-supervised representation learning where the goal is to guarantee the restored image containing sufficient information for semantic recognition. During the optimization procedure, the weights are first assigned to the patch statistics and then gradually lean towards the semantic part. This is to facilitate fast convergence in the early time and then focus on the semantic information that is what we focus on. We shall formulate it as a curriculum learning process.

The proposal leaves two open issues. First, it is unclear whether the adopted semantic proxy (\textit{e.g.}, object detection) is proper for evaluating the recovery quality -- in other words, can we say that $\mathbf{x}'$ is sufficiently similar to $\mathbf{x}$ based on that $\mathbf{x}$ and $\mathbf{x}'$ produce very close object detection results? Second, it is still difficult to adjust $\mathbf{w}\!\left(\mathbf{x},\mathbf{x}'\right)$ within each part, \textit{e.g.}, determining specific patches or objects are more semantically important, so that the pre-trained model learns to focus on these elements. Unfortunately, we did not find any standard to deal with this issue and we conjecture that it is very challenging -- possibly, it is another path that leads to the core difficulty of visual recognition. These two issues are the major obstacle of implementing the algorithm.

Another possible solution lies in a human-in-the-loop evaluation, \textit{i.e.}, asking human labelers to annotate one or few patches or objects on the recovered image that he/she is not satisfying with. Recall the example that a \textit{zebra} stands on the \textit{grass}, the human labeler is expected to focus on the \textit{stripes} of the \textit{zebra} rather than the details of the \textit{grass}. However, this proposal requires human labor during the entire model training procedure, which adds extra costs to the algorithm. By any means, we should encourage more efforts in designing a better evaluation function so that the \textbf{\textit{learning-by-compression}} task can boost visual recognition.

\subsection{Optimization}
\label{proposal:optimization}

The goal of optimization is to improve the tradeoff between $\overline{\mathrm{CR}}$ and $\overline{\mathrm{RQ}}$, as defined in Eqn~\eqref{eqn:criterion}. This is a non-convex optimization and an iterative algorithm is required. In what follows, we assume that a fixed $\overline{\mathrm{CR}}$ is used for all images, though a flexible version can exist by using dynamic networks. We start with a relatively small $\overline{\mathrm{CR}}$, \textit{e.g.}, the configuration of Variant \#1 in Table~\ref{tab:configuration}. An iterative optimization with two steps is performed, with the curriculum that that the evaluation protocol gradually biases towards semantic information (see Section~\ref{proposal:evaluation}) and the compression ratio gradually goes up. The iteration continues until $\overline{\mathrm{RQ}}$ drops below a threshold.

\textbf{In the first step of iteration}, we fix $\overline{\mathrm{CR}}$ and maximize $\overline{\mathrm{RQ}}$. This is done by minimizing the loss function with respect to the parameters of $f\!\left(\cdot\right)$, $g\!\left(\cdot\right)$, $f'\!\left(\cdot\right)$, and $g'\!\left(\cdot\right)$:
\begin{equation}
\label{eqn:loss}
{\mathcal{L}_\mathrm{overall}}={\mathcal{L}_\mathrm{rec}+\lambda_\mathrm{enc}\cdot\mathcal{L}_\mathrm{enc}+\lambda_\mathrm{corr}\cdot\mathcal{L}_\mathrm{corr}},
\end{equation}
where $\mathcal{L}_\mathrm{overall}$ is the overall loss, and $\mathcal{L}_\mathrm{rec}$, $\mathcal{L}_\mathrm{enc}$, and $\mathcal{L}_\mathrm{corr}$ denote denote the separate terms of image recovery, encoding quality, and feature correlation, respectively, and $\lambda_\mathrm{enc}$ and $\lambda_\mathrm{corr}$ are balancing coefficients. $\mathcal{L}_\mathrm{rec}$ is directly related to $\overline{\mathrm{RQ}}$ and involves end-to-end optimization of the encoder-decoder architecture. $\mathcal{L}_\mathrm{enc}$ measures the encoding quality and the form depends on the proxy (\textit{e.g.}, image-level classification or box-level detection losses are added to $f\!\left(\cdot\right)$, mask-level segmentation losses are added to $g\!\left(\cdot\right)$, while self-supervised forms such as the contrastive loss are also possible). $\mathcal{L}_\mathrm{corr}$ measures the correlation between the encoded features, $\left\{\mathbf{z}_l\right\}_{l=1}^L$. Adding this term is to weaken the correlation between features, especially neighboring features, so as to improve the efficiency of visual representation. We emphasize that $\mathcal{L}_\mathrm{rec}$ is the ultimate goal while $\mathcal{L}_\mathrm{enc}$ and $\mathcal{L}_\mathrm{corr}$ are optional -- adding these terms is to stabilize the training procedure. We expect to design an optimization procedure that the balancing coefficients, $\lambda_\mathrm{enc}$ and $\lambda_\mathrm{corr}$, decay gradually.

\textbf{In the second step of iteration}, we maintain $\overline{\mathrm{RQ}}$ and increase $\overline{\mathrm{CR}}$, \textit{e.g.}, altering the configuration from Variant \#1 to Variant \#2 in Table~\ref{tab:configuration}. By increasing the compression ratio, we expect to boost the difficulty of image recovery and facilitate the model towards faster evolution. Technically, we keep the main architecture, $\left\{f_l\!\left(\cdot\right)\right\}_{l=1}^L$ and $\left\{g_l\!\left(\cdot\right)\right\}_{l=1}^L$, unchanged, and truncate the projection layers, $\left\{f_l'\!\left(\cdot\right)\right\}_{l=1}^L$ and $\left\{g_l'\!\left(\cdot\right)\right\}_{l=1}^L$, \textit{i.e.}, the output dimensionality is reduced. To enable a smooth transition from low to high compression ratios, asymmetric dropout can be added to $\left\{f_l'\!\left(\cdot\right)\right\}_{l=1}^L$ and $\left\{g_l'\!\left(\cdot\right)\right\}_{l=1}^L$. For example, if the current dimensionality of $f_l'\!\left(\cdot\right)$ and $g_l'\!\left(\cdot\right)$ is $C_l'$, we can sample an integer ${C_l''}<{C_l'}$ and preserve the first $C_l''$ channels.

We expect the system to be sensitive to the optimizer. In addition, the regular optimization tricks such as degradation-based data augmentation (\textit{e.g.}, Gaussian blur, random noise, RGB-to-grayscale conversion, \textit{etc.}) may help. These hyper-parameters can be tuned beyond the preliminary design.

\subsection{Application to Downstream Tasks}
\label{proposal:application}

The \textbf{\textit{learning-by-compression}} framework offers the ability of data compression, based on which various kinds of downstream vision tasks are made easier. An additional advantage of the proposed framework is that the pre-trained model can be applied to both high-level and low-level vision tasks. This is because the framework contains an encoder and a decoder, and the pre-training objective is not heavily biased towards any particular task.

Applying the pre-trained model for image classification and object detection is straightforward. One needs to extract $f\!\left(\cdot\right)$ from the architecture and equip it with a head, \textit{e.g.}, a linear classification layer or a feature pyramid~\cite{lin2017feature}. It is a little bit different to use the pre-trained model for dense prediction tasks such as semantic segmentation. Both $f\!\left(\cdot\right)$ and $g\!\left(\cdot\right)$ should be inherited, but differently, $g\!\left(\cdot\right)$ needs to be fine-tuned from recovering the original image to predicting the segmentation masks. We expect the fine-tuning procedure to be safe and fast, as pursuing a high compression ratio requires the network to distinguish different semantic regions/instances from each other.

For low-level vision problems, we expect the pre-trained model to unify a wide range of tasks. A typical example is single image super-resolution, where we apply $f\!\left(\cdot\right)$ to achieve compact representation of the degraded image, and then apply the decoder $g\!\left(\cdot\right)$ to recover the high-resolution image. To further improve performance, identity connections can be added between the encoder and decoder, which may require some fine-tuning, yet we still believe the ability of image recovery should make the fine-tuning safe and fast.

The pre-trained model can also be used for image generation, which mostly involves using $f\!\left(\cdot\right)$ for compact representation and fine-tuning $g\!\left(\cdot\right)$. If class information is provided, \textit{i.e.}, conditional image generation, $f\!\left(\cdot\right)$ can also be fine-tuned to integrate the semantic information.

Finally, we discuss the application of the pre-trained model on video understanding. Due to the fewer amount of video data (in terms of samples) and higher computational costs, video understanding is considered more difficult than image understanding. From the perspective of data compression, videos are of higher dimensionality which often incurs heavier redundancy. So, we expect a two-stage pre-training process where the image pre-trained model to serve as the first stage that extracts compact features from each frame, and the second stage follows to further increase the compress ratio of the sequential visual data. For this purpose, sequential models such as LSTM~\cite{hochreiter1997long}, seq2seq~\cite{sutskever2014sequence}, and Transformers~\cite{vaswani2017attention}, can be used.

\section{Related Work, Open Issues, and Discussions}
\label{related}

\subsection{Visual Recognition Requires Fine-scaled Annotations}
\label{related:recognition}

Deep learning~\cite{lecun2015deep} has been dominating the field of visual recognition. The past decade has witnessed a trend of using bigger computational models for better recognition results. A typical example is image classification, where the early models have $10$--$20$ layers~\cite{krizhevsky2012imagenet,simonyan2015very,szegedy2015going} but the leading models nowadays often contain more than $100$ layers~\cite{tan2019efficientnet} or billions of parameters~\cite{goyal2021self}. Training these models requires a large amount of annotations, including image classification~\cite{deng2009imagenet}, object detection and instance segmentation~\cite{everingham2010pascal,lin2014microsoft}, pose estimation~\cite{andriluka20142d,lin2014microsoft}, \textit{etc}.

The above efforts compose of the research path of \textbf{\textit{learning-by-annotation}}, where two clear trends are bigger datasets and finer-level annotations. However, due to the open-set property~\cite{scheirer2012toward,hu2018learning} and compositionality~\cite{zhu2007stochastic,zhu2010part,george2017generative} of real-world concepts, researchers found that the current path may require exponentially amount of data to achieve sufficiently good recognition performance~\cite{yuille2021deep,hinton2021represent}. On the other hand, as the annotation goes to a finer level, the cost is significantly increased, \textit{e.g.}, the average time of annotating a bounding-box is around 2--5 seconds, while it often requires more than one minute to depict the mask of an object. By any means, the \textbf{\textit{learning-by-annotation}} paradigm seems to get close to a plateau and we instead suggest an alternative paradigm, \textbf{\textit{learning-by-compression}}, to learn from unlabeled or weakly-labeled image data.

Though we advocated for bypassing the need of annotations, it remains an open issue whether there exists a finer standard of data annotation to improve the current visual recognition models. This is a challenging topic that mainly involves the definition of objects~\cite{ullman1996high} -- some researchers believed that an object is an individual concept that can be freely removed from other ones, while some others suggested to offer a name or description for each object.

\subsection{Visual Representation Is Driven by Data Compression}
\label{related:compression}

\textbf{\textit{Learning-by-compression}} is not a brand new concept in machine learning. The autoencoder algorithm~\cite{kramer1991nonlinear,hinton2006reducing} was one of the early proposals that the distribution of complex data can be learned by compressing it into low-dimensional vectors and recovering it. There are also variants of autoencoders that regularized the data distribution~\cite{kingma2014auto} or introduced supervision~\cite{gao2015single,le2018supervised} into the model. Some preliminary work was also proposed~\cite{gregor2016towards,graves2018associative} to integrate the autoencoder and semantic representation learning. In the field of cognitive science, there is also a closely related task named minimum length description~\cite{rissanen1978modeling,grunwald2007minimum} that offers a model selection principle, suggesting that the best model can describe data in the shortest form. The relationship of autoencoder and minimum length description was discussed in~\cite{hinton1994autoencoders}. Despite the theory, making use of these principle in deep networks is challenging and preliminary, partly due to the limited ability to offer a reasonable compression bound even under simple data distributions~\cite{blier2018description}. This aligns with our conjecture that pixel-wise distance may incur considerable bias, and designing an evaluation function is the key problem in the future research.

From another perspective, autoencoders and generative adversarial networks~\cite{goodfellow2014generative} were designed to generate complicated data from simple distributions. The GAN inversion problem (\textit{i.e.}, obtaining a low-dimensional representation of an image~\cite{abdal2019image2stylegan,xia2021gan}) can also be considered data compression, yet the community found it difficult not to fine-tune the GAN model during the inversion to fit complicated data distributions~\cite{zhu2020domain}. This inspires us to design a pre-trained model that has the ability of being fine-tuned to fit different kinds of distributions.

\subsection{What is Good for Self-supervised Visual Representation Learning?}
\label{related:learning}

The existing self-supervised visual representation learning algorithms are roughly partitioned into two parts. The first part involves the generative tasks such as autoencoder (see the prior part), inpainting~\cite{pathak2016context}, colorization~\cite{larsson2017colorization}, feature-level prediction~\cite{grill2020bootstrap}, \textit{etc}. Our proposal agrees with them in the need of recovering image contents, but argues that image contents are hierarchical and not all pixels are equally important. The second part involves the discriminative tasks such as spatial relationship prediction~\cite{doersch2015unsupervised,noroozi2016unsupervised,wei2019iterative}, geometry prediction~\cite{gidaris2018unsupervised}, consistency learning~\cite{noroozi2017representation}, contrastive learning~\cite{oord2018representation,chen2020simple,he2020momentum}, \textit{etc}. In particular, researchers discovered that encoding each image into a low-dimensional vector boosts the performance of contrastive learning~\cite{wu2018unsupervised}, which aligns with our opinion that data compression facilitates learning efficient visual representations. Data augmentation plays an important role in improving the difficulty towards better transfer ability~\cite{chen2020simple,chen2020improved}, yet strong data augmentation incurs the risk of feature misalignment~\cite{wang2021dense,xie2021propagate,huo2020heterogeneous}.

A key insight of this paper is to combine generative and discriminative abilities, as either part is incomplete for visual representation learning. In particular, we do not vote for using handcrafted tasks (\textit{e.g.}, contrastive learning or solving jigsaw puzzles) for the discriminative ability, but instead suggest using weak semantic supervision to achieve this goal. This is an alternative way of integrating supervision into self-supervised visual representation learning~\cite{khosla2020supervised,wei2019iterative}. More importantly, our proposal alleviates the conflict between strong data augmentation and consistent feature representation and we expect the learning procedure to be more stable.

\subsection{Data Coverage \textit{vs.} Few-shot Learning Ability}
\label{related:coverage}

\textbf{\textit{Learning-by-compression}}, as other pre-training tasks, pursues a generalized ability of visual representation. This requires large-scale image data to be collected (\textit{e.g.}, Google's JFT-300M, Facebook's Instagram-1B, \textit{etc.}) and a large model with sufficiently many parameters to fit the data. However, this scatters the data distribution and thus increases the difficulty of fine-tuning the pre-trained model into specific domains especially in the few-shot learning scenarios.

We expect the above issue to be a major tradeoff between pre-training and downstream tasks in the future research. A possible solution is to train a super-network or dynamic routing network~\cite{larsson2017fractalnet,huang2018multi} that contains many sub-architectures corresponding to different subsets of training data. In the downstream task, the sub-architecture that is closest to the target domain is extracted and fine-tuned. This idea may also facilitate the development of neural architecture search in which a major obstacle lies in the conflict among sub-architectures~\cite{xie2020weight}.

\subsection{Limitations of the Proposed Framework}
\label{related:limitations}

The performance of \textbf{\textit{learning-by-compression}} depends on two factors, namely, the distribution of the training dataset (semantic labels are considered part of data distribution) and the evaluation function. We conjecture that the paradigm has the following limitations. \textbf{First}, the pre-training task mainly focuses on recovering the original image, and thus it possibly lacks the ability in some particular tasks such as image completion. Meanwhile, it is unclear whether it is easy to fine-tune the pre-trained model to specific data domains, \textit{e.g.}, pre-training on natural image data and fine-tuning on medical image data. \textbf{Second}, the performance of pre-training may be sensitive to the evaluation function, \textit{e.g.}, if an identical weight is added to all pixels, the model may bias towards recovering less meaningful image noise and thus the downstream performance is degraded. \textbf{Third}, the proposed task is designed for visual representation learning. We expect the similar insight to apply to other data modalities such as natural languages and possibly unify multiple modalities together.

\section{Summary}
\label{summary}

This paper describes our idea of a new paradigm for visual representation learning. The motivation is that the current status of visual recognition is far from complete, \textit{i.e.}, recognizing everything that human can recognize, and the traditional \textbf{\textit{learning-by-annotation}} framework has limited potentials to overcome the challenge. Therefore, we propose the \textbf{\textit{learning-by-compression}} framework and suggests a few design principles. We expect the proposal to alter the pursuit of the community from the accuracy-complexity tradeoff to the compression-recovery tradeoff, which we believe is more essential and can incubate pre-trained models for a wide range of vision problems.

\section*{Acknowledgements}

The first author would like to thank Dr. Weichao Qiu, Huiyu Wang, Yanwei Li, Peng Zhou, and Yunjie Tian for instructive discussions.

{\small
\bibliographystyle{abbrv}
\bibliography{refs}
}

\end{document}